\documentclass[sigconf, 10pt]{acmart}


\usepackage{soul}
\usepackage{hyperref}
\usepackage{multirow}
\usepackage[utf8]{inputenc}
\usepackage[ruled, lined, longend, linesnumbered]{algorithm2e}
\usepackage{amsmath,multicol}
\usepackage{subcaption,textcomp,comment}
\usepackage{threeparttable}

\usepackage{booktabs}
\usepackage{multirow}

\usepackage{array}
\newcolumntype{L}[1]{>{\raggedright\let\newline\\\arraybackslash\hspace{0pt}}m{#1}}
\newcolumntype{C}[1]{>{\centering\let\newline\\\arraybackslash\hspace{0pt}}m{#1}}
\newcolumntype{R}[1]{>{\raggedleft\let\newline\\\arraybackslash\hspace{0pt}}m{#1}}

\newcommand{\sys}[0]{\texttt{AUG-FedPrompt}\xspace}

\usepackage{wrapfig}
\usepackage{comment}
\usepackage{colortbl}

\usepackage{hyperref}

\usepackage{color,soul}
\usepackage{graphics}
\usepackage{hyperref}
\usepackage{lipsum}
\usepackage{tikz}
\usepackage{enumitem}	
\usepackage{listings} 

\usepackage{booktabs}	
\usepackage{lipsum}

\usepackage{capt-of}	

\usepackage{todonotes}  
\usepackage{subcaption} 

\definecolor{refkey}{rgb}{0,0,1}
\definecolor{labelkey}{rgb}{0,0,1}

\usepackage{cleveref}
\AtBeginDocument{\DeclareCaptionSubType{lstlisting}}
\crefname{sublstlisting}{listing}{listings}
\Crefname{sublstlisting}{Listing}{Listings}

\usepackage{mathrsfs}	
\usepackage{amsfonts}

\usepackage[export]{adjustbox} 

\usepackage{boldline}					

\usepackage{multirow}

\renewcommand{\paragraph}[1]{\vskip 3pt\noindent\textbf{#1 }}	 

  {\begin{list}{$\bullet$}%
     {\setlength{\parsep}{0pt}%
      \setlength{\topsep}{0pt}%
      \setlength{\itemsep}{2pt}}}%
  {\end{list}}
\newcommand\Noted[1]{} 

\definecolor{darkblue}{rgb}{0.0, 0.0, 0.55}
\definecolor{mygreen}{HTML}{ADFF2F}
\definecolor{mylightgray}{gray}{0.8}

  {\begin{itemize}
	[leftmargin=0cm,
		itemindent=.3cm,
		labelwidth=\itemindent,
		labelsep=0pt,
		parsep=0.0pt,
		topsep=0.5pt,
		itemsep=0.5pt,
		align=left]
  }%
  {\end{itemize}}    

  {\begin{enumerate}
	[leftmargin=.cm,itemindent=.5cm,labelwidth=\itemindent,
		labelsep=0pt,
		parsep=1pt,
		topsep=1pt,
		itemsep=3pt,
		align=left]
  }%
  {\end{enumerate}}    

\makeatletter
\def\@copyrightspace{\relax}
\makeatother

\begin{document}

\title{Towards Practical Few-shot Federated NLP}

\author{Dongqi Cai}
\affiliation{%
  \institution{Beiyou Shenzhen Institute}
  \country{}
}

\author{Yaozong Wu}
\affiliation{%
  \institution{Beiyou Shenzhen Institute}
  \country{}
}

\author{Haitao Yuan}
\affiliation{%
  \institution{Beiyou Shenzhen Institute}
  \country{}
}

\author{Shangguang Wang}
\affiliation{%
  \institution{Beiyou Shenzhen Institute}
  \country{}
}

\author{Felix Xiaozhu Lin}
\affiliation{%
  \institution{University of Virginia}
    \country{}
}

\author{Mengwei Xu}
\affiliation{%
  \institution{Beiyou Shenzhen Institute}
  \country{}
}


\begin{abstract}
Transformer-based pre-trained models have emerged as the predominant solution for natural language processing (NLP).
Fine-tuning such pre-trained models for downstream tasks often requires a considerable amount of labeled private data.
In practice, private data is often distributed across heterogeneous mobile devices and may be prohibited from being uploaded. 
Moreover, well-curated labeled data is often scarce, presenting an additional challenge.
To address these challenges, we first introduce a data generator for federated few-shot learning tasks, which encompasses the quantity and skewness of scarce labeled data in a realistic setting.
Subsequently, we propose \sys, a \textbf{prompt}-based \textbf{fed}erated learning system that exploits abundant unlabeled data for data \textbf{aug}mentation.
Our experiments indicate that \sys can perform on par with full-set fine-tuning with a limited amount of labeled data.
However, such competitive performance comes at a significant system cost.

\end{abstract}


\acmYear{2023}\copyrightyear{2023}
\setcopyright{acmlicensed}
\acmConference[EuroMLSys '23]{3rd Workshop on Machine Learning and Systems}{May 8, 2023}{Rome, Italy}
\acmBooktitle{3rd Workshop on Machine Learning and Systems (EuroMLSys '23), May 8, 2023, Rome, Italy}
\acmPrice{15.00}
\acmDOI{10.1145/3578356.3592575}
\acmISBN{979-8-4007-0084-2/23/05}

\maketitle



\section{Introduction}\label{sec:intro}
\paragraph{Federated NLP}
The development of pre-trained models 
is overwhelming with the rise of BERT~\cite{devlin2018bert}.
Their deployment~\cite{zhang2018deep,shao2019transformer,van2019does,kim2021code, svyatkovskiy2020intellicode} is commonly composed of two-step training:
pre-training and fine-tuning.
Unlike self-supervised pre-training, fine-tuning is supervised, requiring task-specific tremendous labeled data. 
However, the exploitation of private user data is restricted and even prohibited in some cases by several data protection regulations such as GDPR~\cite{voigt2017eu} and CCPA~\cite{pardau2018california}. 
Recently, federate learning (FL)~\cite{mcmahan2017communication,yang2019federated} becomes the de-facto approach to train a model with privacy preserved.
As such, federated NLP (FedNLP)~\cite{lin-etal-2022-fednlp,cai2022autofednlp} is now an important topic towards practical NLP applications.

\paragraph{Problem and challenge}
A key obstacle to practical FedNLP is data labeling. 
It's much more difficult to label data on client devices than on centrally collected data~\cite{xu2021limu, li2017crowdsourced}.
Lack of sufficient labeled data severely limits the practicality and scalability of FedNLP in real-world NLP applications. 
Therefore, it is important to address the issue of few-shot or even zero-shot FedNLP tasks.
There are very few efforts on this topic~\cite{fan2021federated, chen2018federated, huang2022unsupervised}, which still assume a fairly  large number (typically $>$1000 in total) of labels that are uniformly distributed across clients.
However, in practice, the labeled data distribution could be skewed across clients, and such skewness would result in a significant drop in the accuracy according to our experiments in $\S$\ref{sec:related-fl}.

\paragraph{Our solution and contribution}

\noindent (1) To tackle the issue of insufficient and skewness of labeled data, we design a comprehensive data generator as the first step towards simulating the distribution of labeled data for few-shot FedNLP tasks. 
The generator has two meta-parameters: data quantity and skewness, which encompass most, if not all, potential scenarios for practical few-shot FedNLP. 

\noindent (2) To boost the performance of few-shot FedNLP, we design a data-augmented prompt system, namely \sys.
\sys orchestrates prompt learning~\cite{schick2020exploiting} and pseudo labeling~\cite{lee2013pseudo}.
Prompt learning introduces a task description in NLP training.
It helps task-specific fine-tuning achieve high accuracy with very few labeled data samples in FedNLP.
Furthermore, to tackle performance degradation caused by skewed label distribution, \sys leverages enormous and easily accessible unlabeled data for pseudo labeling-based data augmentation.

\noindent (3) Our extensive experiments on four English datasets demonstrate that \sys can achieve a substantial performance gain (25\%--55\% higher accuracy) over the state-of-the-art FedNLP approaches under various few-shot settings.
Augmentation with unlabeled data enhances \sys to perform well with highly skewed labeled distribution across clients.
Overall, \sys can achieve a comparable performance with the state-of-the-art FedNLP approaches with less than 0.1\% labeled data.



\section{Problem setup} \label{sec:related-fl}

\paragraph{Federated NLP Training Procedure}
The two NLP training phases, i.e., pre-training and fine-tuning, require data of disparate natures. 
Pre-training is typically done on public text corpora such as Wikipedia articles, 
while fine-tuning requires domain-specific samples, such as user reviews, messages, or emails. 
For mobile computing, domain-specific samples are gathered from end-users and distributed across mobile devices, while ensuring the protection of privacy. 
To fine-tune models on such private, distributed data, 
federated learning is the de-facto approach~\cite{lin-etal-2022-fednlp,cai2022autofednlp}.
Prior to training, a cloud service distributes a pre-trained model to all client devices.
In a training session targeting a specific NLP task and domain, a cloud service selects multiple mobile devices to participate in training. 
Each device trains a local copy of the model with its private data and sends the model updates to the cloud. 
Upon aggregating the model updates from multiple devices, the cloud sends an updated model to the devices. 
This training procedure is repeated until the model converges.

\begin{figure}[t]
	\centering
	 \includegraphics[width=0.4\textwidth]{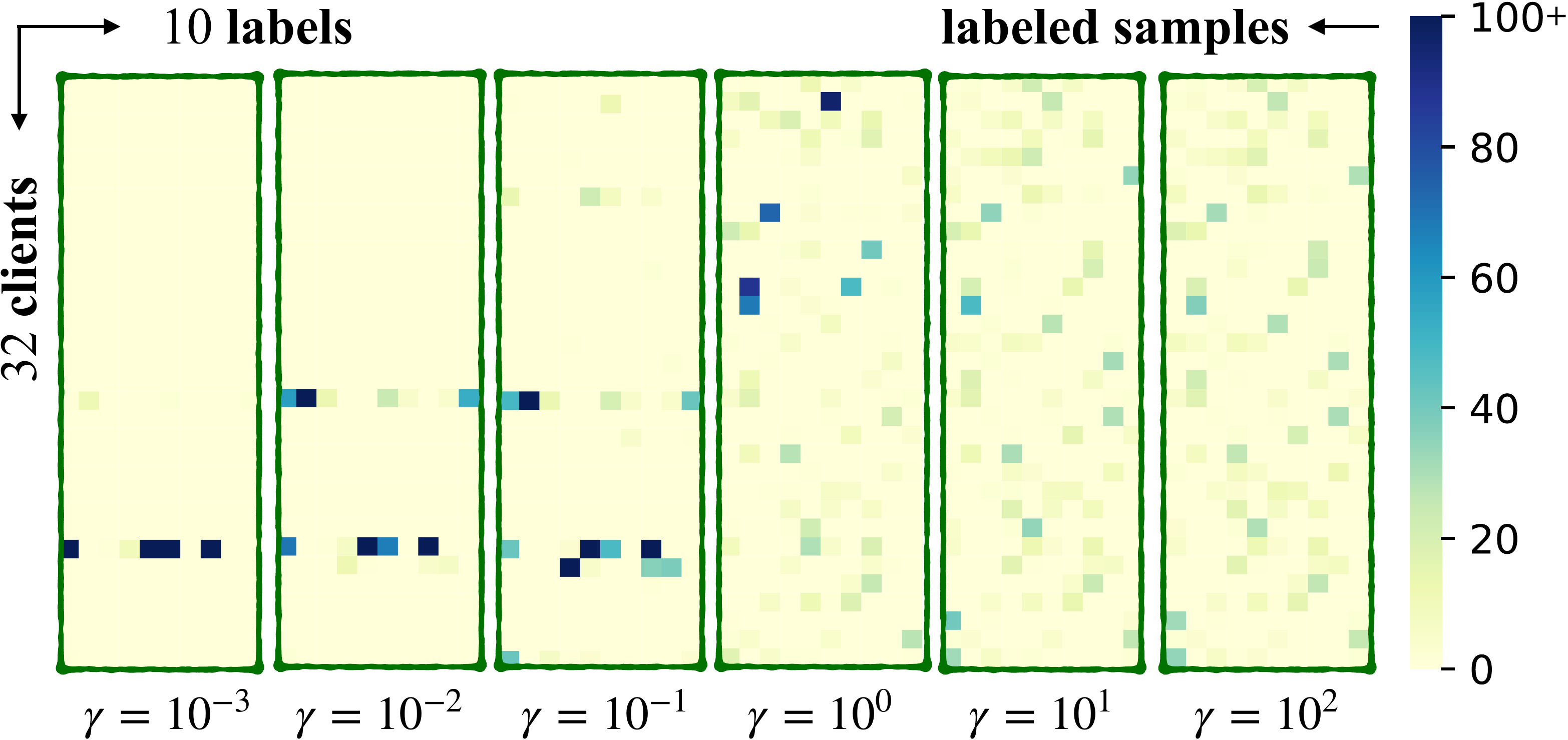}
	\caption{Visualizing the skewness of labeled data on \texttt{YAHOO}~\cite{zhang2015character} with  $n$=1024, $\xi$=32, $\gamma$ being {$10^{n}$, n=-3,-2,..,2}. Each sub-figure is a 32$\times$10 matrix, where 32 is the number of clients and 10 is the number of labels. 
	The intensity of each cell represents the number of labeled samples for a specific label in the client-side local data.} 
	\label{fig:def-gamma}
\end{figure}

\paragraph{Federated few-shot data generator}
Apart from data privacy, lack of sufficient labeled data is a crucial issue and an inherent feature in mobile scenarios.
Alike data feature could be non-independent and identically distributed (non-iid), the scarce labels is not always uniformly distributed in the real world.
Based on the definition of non-iid partition strategies~\cite{li2022federated, lin-etal-2022-fednlp}, we further define the quantity and skewness of labels under federated few-shot learning scenario.

We define a new tuple ($n$, $\gamma$) to represent the practical few-shot learning training data distribution, where $n$ represents the total numbers of labeled data, $\gamma$ represents the skewness of labeled data.

The quantity of labeled data assigned to each client follows a Dirichlet allocation $z$ $\sim$ Dir$_{\xi}$ ($\gamma$), where $\xi$ is the number of clients with labeled data\footnote{$\xi$ could be an optional hyper-parameter to strict the maximum of clients owing labeled data.
In this manuscript, we fix $\xi$ as 32 for simplicity.}.
We can then allocate labeled data from the global labeled dataset to selected clients based on the distribution $z$, with client$_{i}$ being assigned a labeled dataset of size $|\mathcal{T}_{i}|$ = $z_{i}$$n$. 
For example, in Figure~\ref{fig:def-gamma},  we visualize the labeled data skewness on \texttt{Yahoo}~\cite{zhang2015character} with  $n$=1024, $\xi$=32, $\gamma$ being {$10^{n}$, n=-3,-2,..,2}. Each sub-figure is a 32$\times$10 matrix, the intensity of which represents the labeled samples of a particular label. 
When $\gamma$ is small ($10^{-3}$, $10^{-2}$, $10^{-1}$), the labeled data will be skewed distributed, i.e., only few clients own labeled data; when $\gamma$=$10^{2}$, labeled data is nearly uniformly distributed on all clients.

\textbf{Performance degradation under skewed distribution}
\begin{figure}[t]
    \centering
    \includegraphics[width=0.234\textwidth]{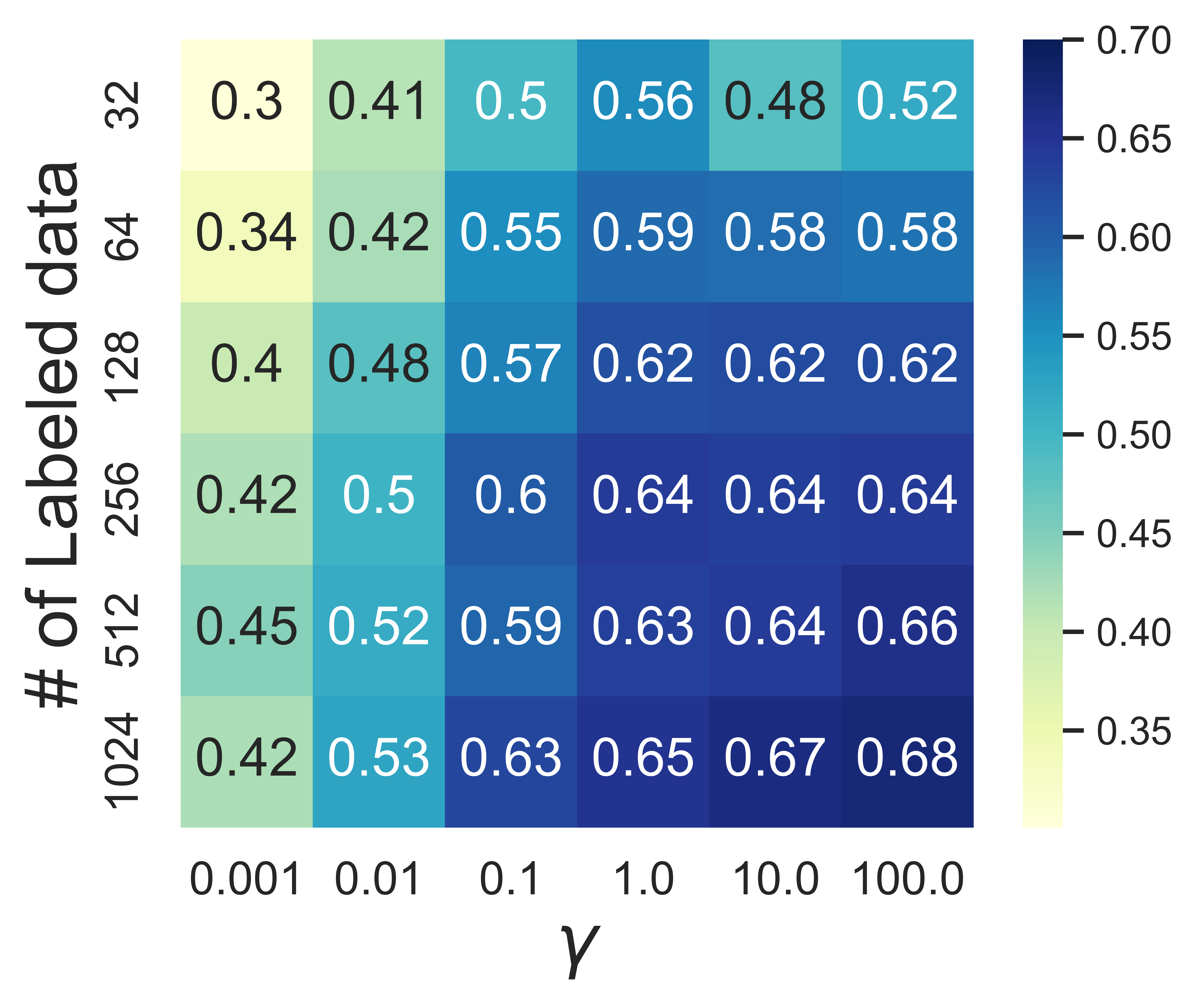}
    \vspace{-4pt}
    \caption{Average accuracy of federated few-shot learning under different data quantity and skewness. When skewness $\gamma$ grows larger, labeled data will be more uniformly distributed, and vice versa. Dataset: \texttt{YAHOO}~\cite{zhang2015character}.}    
    \label{fig:eval-sparsity-new}
    
\end{figure}

In Figure~\ref{fig:eval-sparsity-new}, we present the impact of label skewness on federated few-shot learning.
We observe that as $\gamma$ decreases, i.e., the labeled data becomes more skewed, the convergence performance of the model degrades.
For example, when labeled data points are 1024, uniform distribution ($\gamma$=100) will be 26\% better than skewed distribution ($\gamma$=0.001).
The rationale behind this phenomenon is that under common non-iid data distribution, individual clients tends to possess more specific data features.
The labels concentrated on certain clients results in a skewed feature distribution of training data. 
This bias can lead to unfairness, as the aggregated model may favor certain labels over others, resulting in a significant drop in convergence accuracy.
We provide a more detailed analysis of this phenomenon in Section~\ref{sec:eval-gamma}.

\section{System design} \label{sec:related-fsl}
We propose \sys as a solution to address the challenges posed by data privacy concerns and label scarcity. 
\sys leverages large amounts of unlabeled data via the federated orchestration of pseudo labeling and prompt learning. 
\sys fine-tunes the pre-trained model through prompt learning on client devices. 
After local training and federated aggregation, \sys 
makes inference on unlabeled training data, from which high-confident results, i.e., pseudo labels~\cite{lee2013pseudo,arazo2020pseudo,bengio2009curriculum} are selected for subsequent training.

\begin{figure}[t]
	\centering
	 \includegraphics[width=0.49\textwidth]{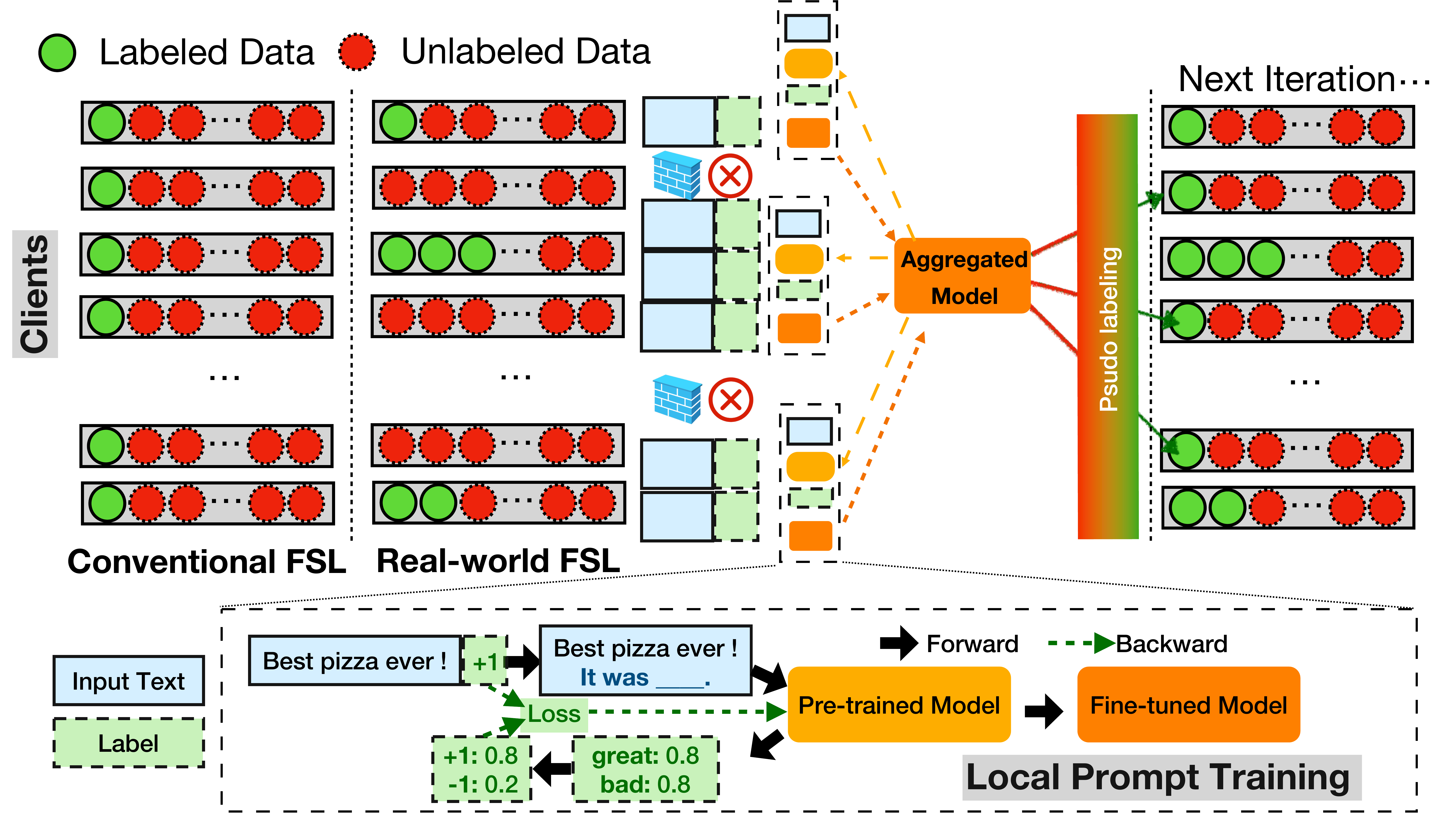}
	\caption{Workflow of \sys. 
	} 
	\label{fig:into-framework}
\end{figure}
We describe the training workflow of \sys in Figure~\ref{fig:into-framework}.
A public pre-trained transformer-based language model $M$ is transferred to chosen clients.
We assume that each client has access to a tiny training set $\mathcal{T}$ (typically $<$ 10) and a much larger set of unlabeled examples $\mathcal{D}$ (typically $>$ 1000).

For local prompt training, we annotate $T$ as the vocabulary of model $M$, $\_\_ \in T$ as the mask token and $T^*$ as the set of all token sequences.
To clarify, $T$ is composed of tokens representing labels description and $T^*$ is composed of tokens representing input text, which is a larger corpus.
The sequence of input phrases is $\mathbf{x}=\left(s_1, \ldots, s_k\right)$ where $s_i \in T^*$.
The pattern-verbalizer pair $\mathbf{p}$ includes: 
1) a \textit{pattern} $P: X \rightarrow T^*$ maps inputs $x$ to a cloze question containing a single mask; 
2) a \textit{verbalizer} $v: Y \rightarrow T$ maps each output $y$ to a single token representing its task-specific meaning in the pattern.

The purpose of local prompt training is to derive the probability that y is the correct output of x from the probability that v(y) is the most likely token at the masked position in $P(x)$.
Based on this rationale, we define the conditional probability distribution $s_{\mathbf{p}}$ of $y$ given $x$ as:

\begin{equation}
    s_{\mathbf{p}}(y \mid x)=\frac{\exp q_{\mathbf{p}}(y \mid x)}{\sum_{y^{\prime} \in Y} \exp q_{\mathbf{p}}\left(y^{\prime} \mid x\right)} \label{eq:pet-few}
\end{equation}
where $q_{\mathbf{p}}(y \mid x)=M(v(y) \mid P(x))$ is the probability that $M$ assigns to $v(y)$ in sequence $P(x)$.

For client-side fine-tuning, the pre-trained model 
$M$ is fine-tuned on local labeled data $(x, y)$ by minimizing the cross-entropy  between $s_{\mathbf{p}}(y \mid x)$ and y.
For server-side aggregation, in each iteration $i$, client $k$ sends its updated model $M^{i}_{k}$ to the cloud for aggregation using FedAVG algorithm~\cite{mcmahan2017communication}; the aggregated model is denoted as $M^{i}$.

For data augmentation, $M^{i}$ is distributed to clients with large amount of unlabeled data for pseudo labeling.
Each unlabeled example $\hat{x} \in D$ is labeled with pseudo label $\hat{y}$ based on $s_{\mathbf{p}}(\hat{y} \mid \hat{x})$.
The pseudo-labeled dataset then is utilized for fine-tuning the client-side model in the subsequent iteration.

The resulting pseudo-labeled dataset could consist of enormous samples with wrong labels.
Directly involving them in the next training iteration will poison the foundation model, which makes it could be even worse than purely using the limited labeled data.
To address this issue, we propose two techniques to filter out those wrong samples and remain the purity of augment dataset:
1) Filtering by model capacity: we eliminate those models with low model capacity, i.e., those that perform poorly on validation datasets.
2) Filtering by confidence: we remove samples with low confidence, i.e., those with a probability of the most likely label lower than a threshold.
Both capacity and confidence are hyper-parameters that can be tuned flexibly depending on particular tasks or datasets.

\section{Preliminary Experiments} \label{sec:eval}

In this section, we evaluate the performance of \sys across data scales.
\sys significantly outperforms naive federated fine-tuning.
It could perform on par with full-set training while saving up to 99.9\% labeled data.
Apart from data efficiency, \sys shows great robustness under various practical few-shot scenario regardless of skewed or uniform label distribution. 

\subsection{Experiment Setup} \label{sec:eval-setup}
\begin{table}[t]
    \centering
    \footnotesize
    \begin{tabular}{@{}cc|cc@{}}
    \toprule
    \textbf{Dataset} & \textbf{Prompt}   & \textbf{Train}  & \textbf{Test}    \\ \midrule
    \texttt{AGNEWS}~\cite{zhang2015character}  & a ( \_\_\_\_ ) b    & 120,000    & 7.600  \\
    \texttt{MNLI}~\cite{williams2017broad}    & “a” ? ‖ \_\_\_\_, “b”   & 392,702     & 9,815  \\
    \texttt{YAHOO}~\cite{zhang2015character}   & [ Category: ] a \_\_\_\_ b      & 1,400,000    & 60,000  \\
    \texttt{YELP-F}~\cite{zhang2015character}    & It was \_\_\_\_. a & 650,000       & 50,000   \\ \bottomrule
    \end{tabular}
    \caption{Evaluation datasets. Each dataset is distributed to 1000 clients. Label quantity of each class follows the non-iid label distribution in~\cite{lin-etal-2022-fednlp} where $\alpha=1$.}
    \label{tab:eval-datasets}
\end{table}
\begin{figure*}[t]
    \centering
    \begin{minipage}[b]{1\textwidth}
        \begin{minipage}[b]{0.24\textwidth}
            \includegraphics[width=1\textwidth]{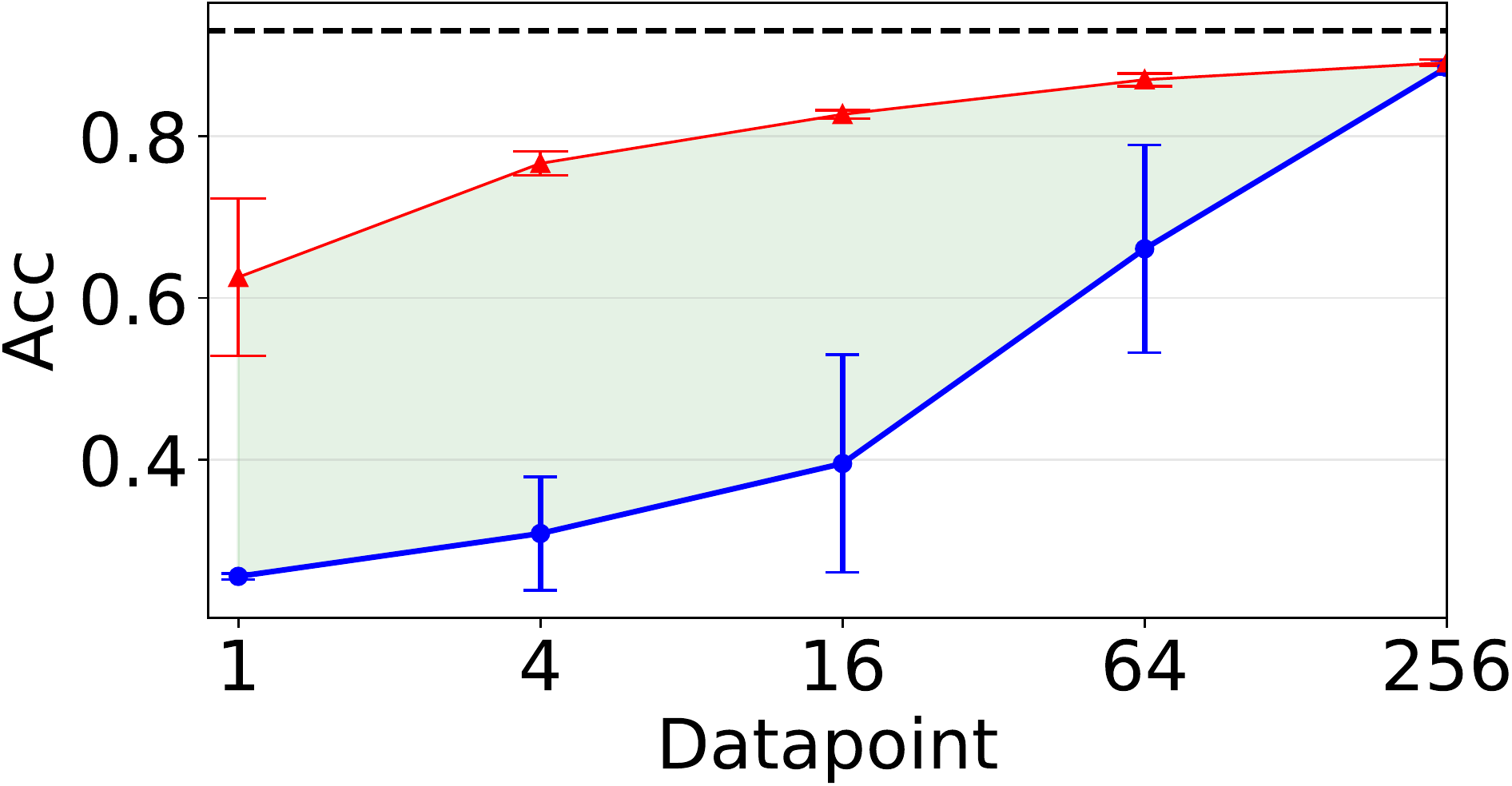}
            \subcaption{\texttt{AGNEWS}}
        \end{minipage}
        ~
        \begin{minipage}[b]{0.24\textwidth}
            \includegraphics[width=1.83\textwidth]{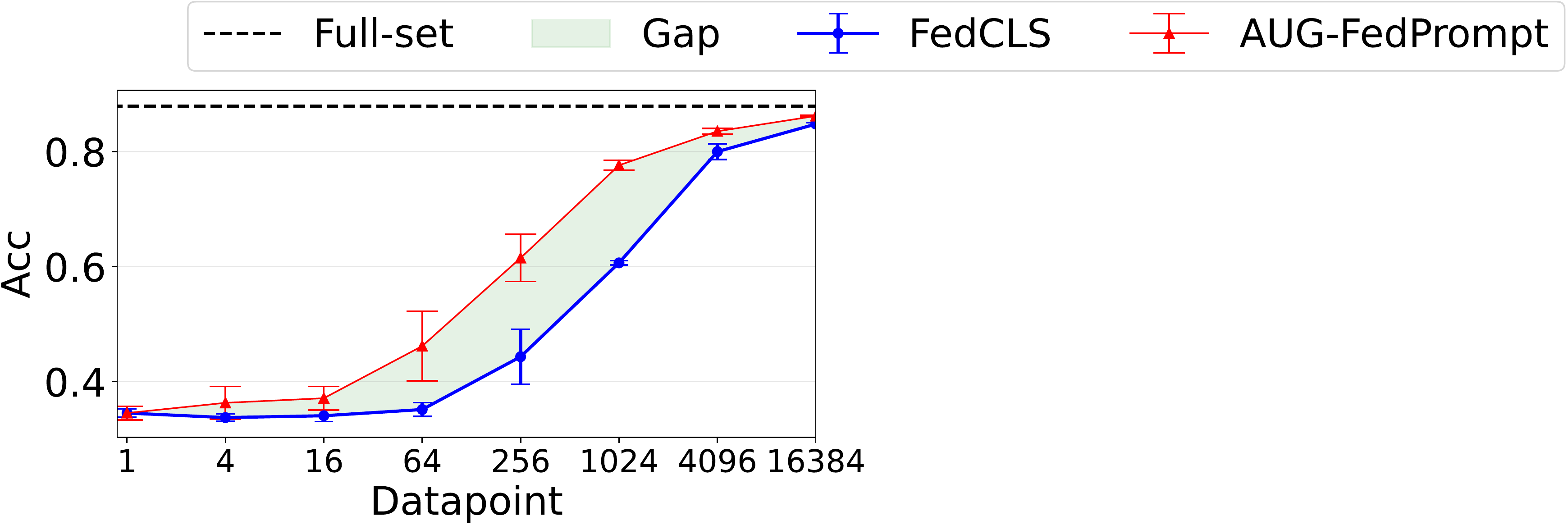}
            \subcaption{\texttt{MNLI}}
        \end{minipage}
        ~
        \begin{minipage}[b]{0.24\textwidth}
            \includegraphics[width=1\textwidth]{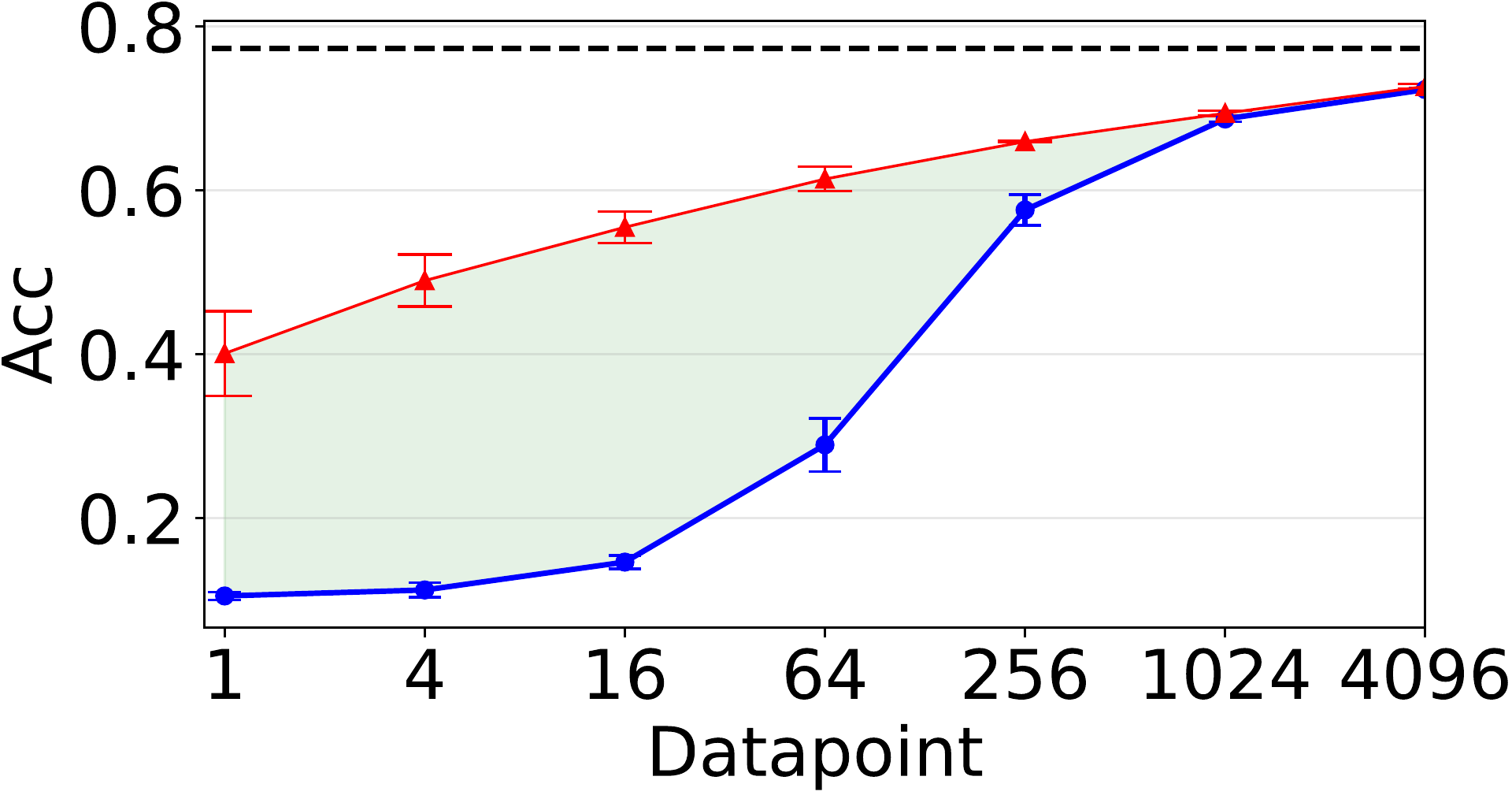}
            \subcaption{\texttt{YAHOO}}
        \end{minipage}
        ~
        \begin{minipage}[b]{0.24\textwidth}
            \includegraphics[width=1\textwidth]{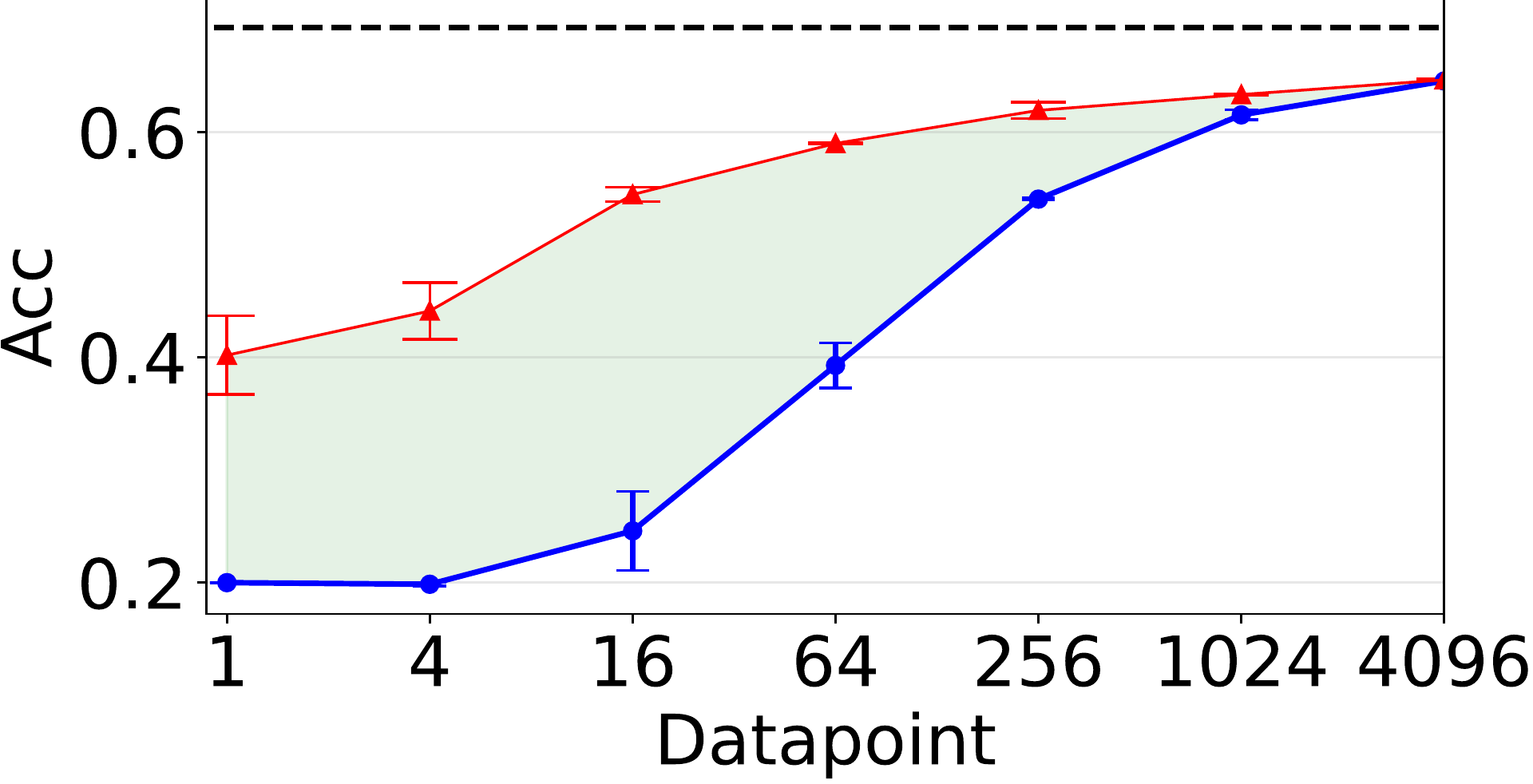}
            \subcaption{\texttt{YELP-F} }
        \end{minipage}
    \end{minipage}
    
    \caption{Average accuracy and standard deviation for AUG-FedPrompt across data scales. FedCLS stands for the vanilla federated fine-tuning. Full-set stands for fine-tuning on the full labeled data.}
    \label{fig:eval-performance-datapoint}
\end{figure*}

\paragraph{Dataset and models} 
We perform our evaluation on four  English datasets and manually designed prompts\footnote{We try 6, 2, 6, 4 different prompts for each datasets separately and report the chosen one that performs best.
The verbalizers are the same as previous literature~\cite{schick2020exploiting}.},
 detailed information is shown in Table~\ref{tab:eval-datasets}.
(1) \texttt{AGNEWS}~\cite{zhang2015character} is a news classification dataset. Given headline and text body, news needs to be classified as one of the four categories: World, Sports, Business or Science/Tech.
(2) \texttt{MNLI}~\cite{williams2017broad} is a sentence understanding dataset. Given text pairs x = (a, b), the task is to find out whether a implies b, a and b contradict each other or neither.
(3) \texttt{YELP Review Full (YELP-F)}~\cite{zhang2015character} is a restaurant rating dataset. Given a customer's review, text should be estimated on a 1-5 star scale.
(4) \texttt{YAHOO}~\cite{zhang2015character} is a text classification dataset. Given a question and an answer, one of ten possible categories needs to be assigned.
All experiments are conducted using the same pre-trained model, RoBERTa-large (355M parameters)~\cite{liu2019roberta}, which we load from the transformers~\cite{wolf2019huggingface} library.

\paragraph{Hyper-parameters} 
In line with previous observations~\cite{scao2021many}, few-shot fine-tuning performance varies across chosen labeled data considerably. 
We run every experiment 3 times in order to reduce variance.
Unless otherwise stated, we use the recommended set of hyper-parameters from previous work~\cite{schick2020exploiting}: 
mini-batch size as 4; local training iteration as 1; learning rate as 10$^{-5}$; max sequence length as 256.
For pseudo labeling, we filter out those aggregated models performing worse than the zero-shot model and those pseudo-labeled data with confidence lower than 0.9.
For the FL configurations at the server side, we follow the prior FedNLP literature~\cite{cai2022autofednlp,lin-etal-2022-fednlp} to select 5 participants for each training round by default.
The fine-tuned models will be collected in the central server and aggregated through \textit{FedAvg} algorithm~\cite{mcmahan2017communication}.




\subsection{Performance across Data Scales}  \label{sec:eval-performance}

\textbf{\sys enjoys a substantial advantage on each task.}
As shown in Figure~\ref{fig:eval-performance-datapoint}, we compare our \sys performance with FedCLS, i.e., the vanilla federated fine-tuning where a generic classifier layer inserted after pre-trained models is fine-tuned.
Highlighted region shows the accuracy gap between \sys and FedCLS. 
There are up to 50\%, 25\%, 55\%, 38\% accuracy improvement separately for 4 datasets.
Both approaches improve with more labeled data, but \sys remains better by a varying amount. 
\sys reaches 99\% relative performance of full-set with 90\% less training data compared to full-set federated training.
\sys shows a strong zero-shot inferring capability, i.e., without task-specific fine-tuning, expect for \texttt{MNLI} dataset. 
\texttt{MNLI} dataset may need more labeled data to make full use of the prompt to the pre-trained models.
For a usable accuracy, i.e., 90\% relative performance of full-set training accuracy, \sys only needs 64, 256, 256 in total for \texttt{AGNEWS}, \texttt{YAHOO} and \texttt{YELP-F}, saving up to 99.9\% training data compared to full-set federated fine-tuning.
Please note that 64 is the total number of labels across all clients, not per client.

\subsection{Impact of Data Augmentation}  \label{sec:eval-gamma}
\begin{table}[t]
    \centering
    \scriptsize
    \begin{tabular}{@{}cc|cccc@{}}
    
    \toprule
    \multicolumn{2}{c|}{\textbf{Dataset}}            & \textbf{AGNEWS} & \textbf{MNLI} & \textbf{YAHOO} & \textbf{YELP-F} \\ \midrule
    \multirow{2}{*}{\textbf{Uniform}} & FedCLS   & 66.1$\pm$12.8    & 60.1$\pm$0.4 & 57.6$\pm$1.9  & 54.0$\pm$0.1 \\
                            & FedPrompt     & \textbf{87.0}$\pm$0.8   & \textbf{77.6}$\pm$0.8 & \textbf{66.0}$\pm$0.1  & \textbf{61.9}$\pm$0.7 \\
                            \midrule
    \multirow{3}{*}{\textbf{Skewed}}  & FedCLS   & 64.8$\pm$3.1   & 37.7$\pm$5.6 & 24.4$\pm$10.3  & 38.3$\pm$8.8 \\
                            & FedPrompt     & 68.4$\pm$2.4   & 42.4$\pm$5.8 & 41.8$\pm$4.3  & 51.2$\pm$1.8 \\
                            & w/ augment & \textbf{90.2}$\pm$0.5   & \textbf{75.7}$\pm$1.2 & \textbf{66.9}$\pm$1.1  & \textbf{58.2}$\pm$2.4 \\ \bottomrule
    \end{tabular}
    \caption{AUG-FedPrompt enhances performance under different few-shot learning settings. 
    FedPrompt stands for AUG-FedPrompt without unlabeled data augmentation.
    Datapoint: 64 for AGNEWS, 1024 for MNLI, 256 for YHAOO and YELP-F. }
    \label{tab:eval-softlabel}
\end{table}
\textbf{\sys enhances FedPrompt performance when labeled data is skewed distributed.}
As shown in Table~\ref{tab:eval-softlabel}, FedPrompt, i.e., \sys without data augment shows competitive performance when labeled data is uniformly distributed on clients.
While skewed distribution of labeled data will hurt FedPrompt performance significantly.
For example, FedPrompt performance degrades to 41.8\% on \texttt{YHAOO} when 256 labeled data is skewed distributed on 32 clients.
Considering that skewed distribution is common in real-world, we integrate \sys with data augmentation to mitigate the performance degradation.

It is important to recall that prompts learning introduces a task description in NLP training.
Prompt helps task-specific fine-tuning perform well even with few labeled training data.
This rationale paves the way for the efficiency of pseudo labeling; it helps to label more data correctly at the early stage of training.
Together with our confidence filter for pseudo-labeling, \sys makes pseudo-labeled data seldom hurt.
For example, we annotate 100 unlabeled data on each client involved in per round for \texttt{AGNEWS}.
In the first three rounds, the average ratio of correctly labeled data by pseudo-labeling on unlabeled data is 92.5\%.
The inference accuracy will further increase along with the FL training moves on, reaching 95.3\% at the convergence round.
Those `nail' data, about 5 out of 100 in total, is hard to be correctly annotated and filtered out.
Fortunately, we observe that they do not affect the model convergence as shown in Table~\ref{tab:eval-softlabel}.
After pseudo labeling, \sys performs on par with full-set fine-tuning and greatly outperforms vanilla few-shot fine-tuning, reaching a usable accuracy with scarce labeled data.

\section{System Cost}\label{sec:cost}
\begin{table}[]
\resizebox{\columnwidth}{!}{%
\begin{tabular}{@{}ll@{}}
\toprule
\textbf{Challenges}               & \textbf{Possible Solutions}            \\ \midrule
Huge training latency    & Model structure optimization~\cite{sanh2019distilbert, lan2019albert}. \\
\rowcolor[HTML]{EFEFEF} 
Large memory requirement & Rematerialization~\cite{chen2016training, wang2022melon}, paging~\cite{peng2020capuchin}.    \\
Excessive inference for pseudo labeling & Pacing~\cite{cascante2021curriculum,bengio2009curriculum}, early-exit~\cite{zhou2020bert, laskaridis2021adaptive}.      \\
\rowcolor[HTML]{EFEFEF} 
High communication cost                 & Quantization~\cite{wu2018error, abdelmoniem2021towards}, sparsity~\cite{li2021hermes,frankle2018lottery}. \\ \bottomrule
\end{tabular}%
}
\caption{Challenges and possible solutions.}
\label{tab:cost-challenges}
\end{table}
There is no free lunch for the performance improvement of \sys.
The orchestrating of pseudo labeling and prompt learning results in promising few-shot performance, but it also incurs a non-trivial system cost. 
In this section, we discuss the necessity of large models for \sys, as well as the associated system cost.
Challenges and possible solutions are concluded in Table~\ref{tab:cost-challenges}.

\begin{figure}[t]
	\centering
        \includegraphics[width=0.5\textwidth]{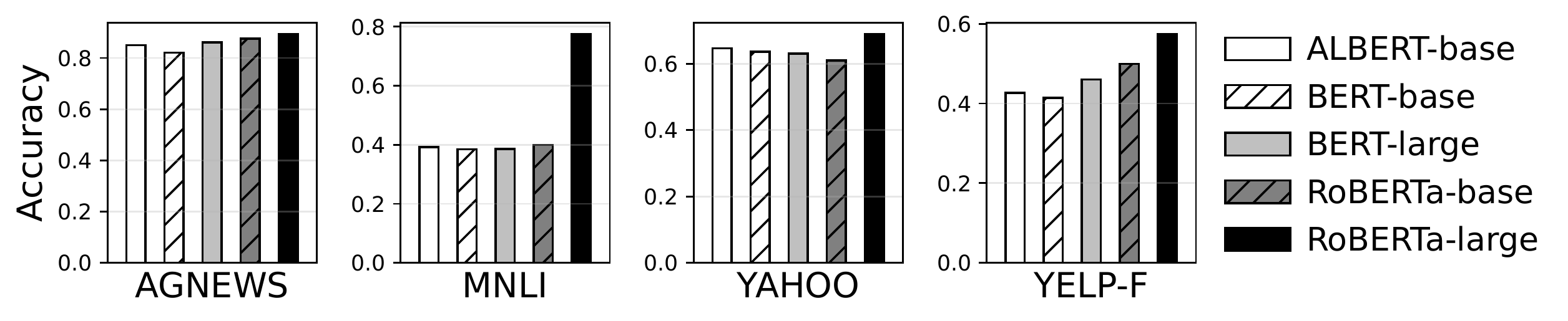}

	\caption{\sys convergence performance with different models and datasets. 0.1\% labeled data uniformly distributed in 32 clients.} 

	\label{fig:eval-model}
\end{figure}

To begin with, we conduct experiments to evaluate the performance of \sys on various foundation models.
As demonstrated in Figure~\ref{fig:eval-model}, RoBERTa-large outperforms all other models across all four datasets, particularly MNLI and YELP-F, where it shows a significant improvement (up to 38.2\%).
In contrast, BERT-large, despite having similar parameters to RoBERTa-large, performed poorly. Interestingly, certain small models, e.g. ALBERT-base~\cite{lan2019albert}, which is optimized from BERT-base achieved superior results compared to the standard BERT-base model, despite containing only 10.7\% of the parameters. These findings suggest that large models can help augment the few-shot learning abilities of \sys, and that model structure optimization shows promise in making \sys a more practical solution.

The excellent performance of RoBERTa-large aligns with previous research~\cite{schick2020exploiting, scao2021many, schick2022true}, highlighting the need for large-scale foundational models to fully leverage prompt learning. 
However, despite its merits, the model's high memory usage and latency cannot be overlooked.
As shown in Table~\ref{tab:cost-model}, even on a powerful GPU-embedded edge device like NVIDIA TX2~\cite{tx2}, training RoBERTa-large leads to long latency (about 8.1s per batch).
Moreover, during training, our testbed device, which has only 8GB of RAM, ran out of memory during training.
Because the peak memory usage of RoBERTa-large fine-tuning  is over 10GB\footnote{Tested on a central server.}.
\begin{table}[]
    \resizebox{\columnwidth}{!}{%
    \begin{tabular}{@{}cccccc@{}}
    \toprule
    \textbf{Model} &
      \textbf{\begin{tabular}[c]{@{}c@{}}ALBERT-base\\ \cite{lan2019albert}\end{tabular}} &
      \textbf{\begin{tabular}[c]{@{}c@{}}BERT-base\\ \cite{devlin2018bert}\end{tabular}} &
      \textbf{\begin{tabular}[c]{@{}c@{}}BERT-large\\  \cite{devlin2018bert}\end{tabular}} &
      \textbf{\begin{tabular}[c]{@{}c@{}}RoBERTa-base\\ \cite{liu2019roberta}\end{tabular}} &
      \textbf{\begin{tabular}[c]{@{}c@{}}RoBERTa-large\\ \cite{liu2019roberta}\end{tabular}} \\ \midrule
    \multicolumn{1}{c|}{\textbf{Memory (GB)}} & 3.7  & 5.4   & OOM (9.8) & 5.8   & OOM (10.4) \\
    \multicolumn{1}{c|}{\textbf{Latency (s)}} & 1.4  & 1.9   & $\sim$7.8       & 2.1   & $\sim$8.1        \\
    \multicolumn{1}{c|}{\textbf{Param. (M)}}  & 11.7 & 109.5 & 334.9     & 124.6 & 355.3      \\ \bottomrule
    \end{tabular}%
    }
    \caption{System cost of different NLP models. Tested on NVIDIA TX2. Batch size: 4.}
    \label{tab:cost-model}
    \end{table}

Apart from local prompt training, a mobile client need to perform inference on \textit{all} of its unlabeled data to generate pseudo labels. 
However, most of this inference is ultimately unnecessary, as only a small fraction (the most confident) of pseudo labels will be selected for subsequent training. 
As a result, the inference process dominates the total delay due to the large volume of unlabeled data that needs to be processed. 
According to our measurements, this process accounts for up to 87.4\% of the total computation time.
Keeping a balanced pace between training and labeling is crucial to reduce those redundant inference.

In addition, it should be noted that the overall resource cost of \sys system should be extremely higher, let alone long heavy-duty computing. 
The reason for this is the need to transfer the entire model, which can be several GBs in size, in a federated learning scenario. 
As the size of the model increases, so too does the amount of data that needs to be transferred, leading to higher communication costs.
This can be particularly problematic in settings with limited network bandwidth, such as mobile devices, where large network traffic can significantly impact system performance~\cite{wu2018error, xu2020client, reisizadeh2020straggler, wang2021device}.

\section{Conclusions and Future Work}\label{sec:conclusions}

This manuscript explores a crucial but less explored issue: data labels can be scarce in federated learning.
We provide a comprehensive definition of a data generator for federated few-shot learning tasks and demonstrate that the lack and skewness of labeled data can significantly degrade federated learning convergence performance. 
To mitigate this issue, we propose \sys, a novel federated few-shot learning system that orchestrates prompt learning and pseudo labeling.
\sys shows competitive performance under various federated few-shot learning settings, requiring less than 0.1\% data to be manually labeled.

In conclusion, our experiments have demonstrated the impressive few-shot performance of \sys when used with large-scale pre-trained models. 
However, fine-tuning these `behemoths' can be extremely resource-intensive, requiring significant computational power and memory. 
Additionally, the communication of large model parameters can consume a considerable amount of bandwidth. 
Future work will focus on the development of an optimized system solution for \sys to enhance its resource efficiency.
\section*{Acknowledgments}

This research was supported by National Key Research and Development Program of China \#2020YFB1805500, the Fundamental Research Funds for the Central Universities, and NSFC \#62032003, \#61922017, \#61921003. 
Mengwei Xu was partly supported by NSFC \#62102045, Beijing Nova Program \#Z211100002121118, and Young Elite Scientists Sponsorship Program by CAST \#2021QNRC001. 
The authors thank the anonymous reviewers for their insightful feedback.

\balance
\bibliographystyle{IEEEbib}
\bibliography{bib/ref-mwx,bib/nlp,bib/ref-cdq,bib/autofednlp}

\end{document}